\documentclass[10pt,twocolumn,letterpaper]{article}

\usepackage{cvpr}
\usepackage{times}
\usepackage{epsfig}
\usepackage{graphicx}
\usepackage{amsmath}
\usepackage{amssymb}
\usepackage{subfigure}

\usepackage{amsmath,amsfonts,bm,amssymb}

\def\eqref#1{equation~\ref{#1}}

\def\1{\bm{1}}

\def\rvx{{\mathbf{x}}}

\DeclareMathAlphabet{\mathsfit}{\encodingdefault}{\sfdefault}{m}{sl}
\SetMathAlphabet{\mathsfit}{bold}{\encodingdefault}{\sfdefault}{bx}{n}

\def\bmu{{\boldsymbol{\mu}}}
\def\bgamma{{\boldsymbol{\gamma}}}
\def\bomega{{\boldsymbol{\omega}}}
\def\bbeta{{\boldsymbol{\beta}}}

\usepackage{color}
\usepackage{url}
\usepackage{amsfonts}
\usepackage{multirow}
\newcommand{\comment}[1]{}

\definecolor{Crimson}{rgb}{0.86, 0.08, 0.24}
\definecolor{DarkGreen}{rgb}{0.00, 0.40, 0.00}
\definecolor{RoyalBlue}{rgb}{0.15, 0.25, 0.54}
\definecolor{DarkCyan}{rgb}{0.0, 0.54, 0.54}
\ifx \submission \undefined

\newcommand{\djc}[1]{\textcolor{DarkGreen}{#1}}

\else

\newcommand{\djc}[1]{{#1}}

\fi

\usepackage[pagebackref=true,breaklinks=true,letterpaper=true,colorlinks,bookmarks=false]{hyperref}

\cvprfinalcopy 

\def\cvprPaperID{3013} 

\ifcvprfinal\fi
\begin{document}

\title{Instance-Level Meta Normalization}

\author{Songhao Jia\\
National Tsing Hua University\\
{\tt\small gasoonjia@gapp.nthu.edu.tw}
\and
Ding-Jie Chen\\
Academia Sinica\\
{\tt\small djchen.tw@gmail.com}
\and
Hwann-Tzong Chen\\
National Tsing Hua University\\
{\tt\small htchen@cs.nthu.edu.tw}
}

\maketitle

\begin{abstract}
This paper presents a normalization mechanism called Instance-Level Meta Normalization (ILM~Norm) to address a learning-to-normalize problem. ILM~Norm learns to predict the normalization parameters via both the feature feed-forward and the gradient back-propagation paths.
ILM~Norm provides a meta normalization mechanism and has several good properties. It can be easily plugged into existing instance-level normalization schemes such as Instance Normalization, Layer Normalization, or Group Normalization. ILM~Norm normalizes each instance individually and therefore maintains high performance even when small mini-batch is used. The experimental results show that ILM~Norm well adapts to different network architectures and tasks, and it consistently improves the performance of the original models. The code is available at  \url{https://github.com/Gasoonjia/ILM-Norm}. 

\comment{Normalization is a key technique to improve the training of deep neural networks. However, the existing normalization methods often merely rely on the back-propagation processes to learn the rescaling parameters. That is, the existing methods treat all \emph{different} input features under the \emph{same} distribution, which may limit the feature expressiveness of the normalization module. We present Associate Normalization (AssocNorm) to overcome the above limitation. AssocNorm extracts the useful information from input features and associates them with rescaling parameters predicted by an auto-encoder-like neural network. Therefore, AssocNorm learns the rescaling parameters via both the back-propagation and the association with input features. Furthermore, AssocNorm normalizes the features of each example individually, so the accuracy is relatively stable for different batch sizes. The experimental results show that AssocNorm outperforms the existing normalization methods on several benchmark datasets under various hyper-parameter settings. 
}
\end{abstract}

\section{Introduction}
 The mechanism of normalization plays a key role in deep learning. Various normalization strategies have been presented to show their effectiveness in stabilizing the gradient propagation. In practice, a normalization mechanism aims to normalize the output of a given layer such that the vanishing gradient problem can be suppressed and hence to reduce the oscillation in the output distribution. With appropriate normalization, a deep network would be able to improve the training speed and the generalization capability. 

A typical normalization mechanism contains two stages: \emph{standardization} and \emph{rescaling}. The standardization stage regularizes an input tensor $\rvx$ of feature maps with its mean $\bmu$ and variance $\bgamma$ by
\begin{equation}
	\rvx_s = \frac{\rvx - \bmu}{\sqrt{\bgamma + \epsilon}} \,,
\end{equation}
\noindent where $\rvx_s$ is a standardized input feature tensor. 
At the rescaling stage, the standardized feature tensor $\rvx_s$ is rescaled by a learned weight $\bomega$ and a bias $\bbeta$ to recover the statistics of the features vanished in the standardization stage
by 
\begin{equation}
	\rvx_n = \bomega \ast \rvx_s  + \bbeta \,,
    \label{eq:rescaling}	
\end{equation}  
where $\rvx_n$ is the final output of the entire normalization process. 


\begin{figure}[!t]
  \centering
  \includegraphics[width=1\linewidth]{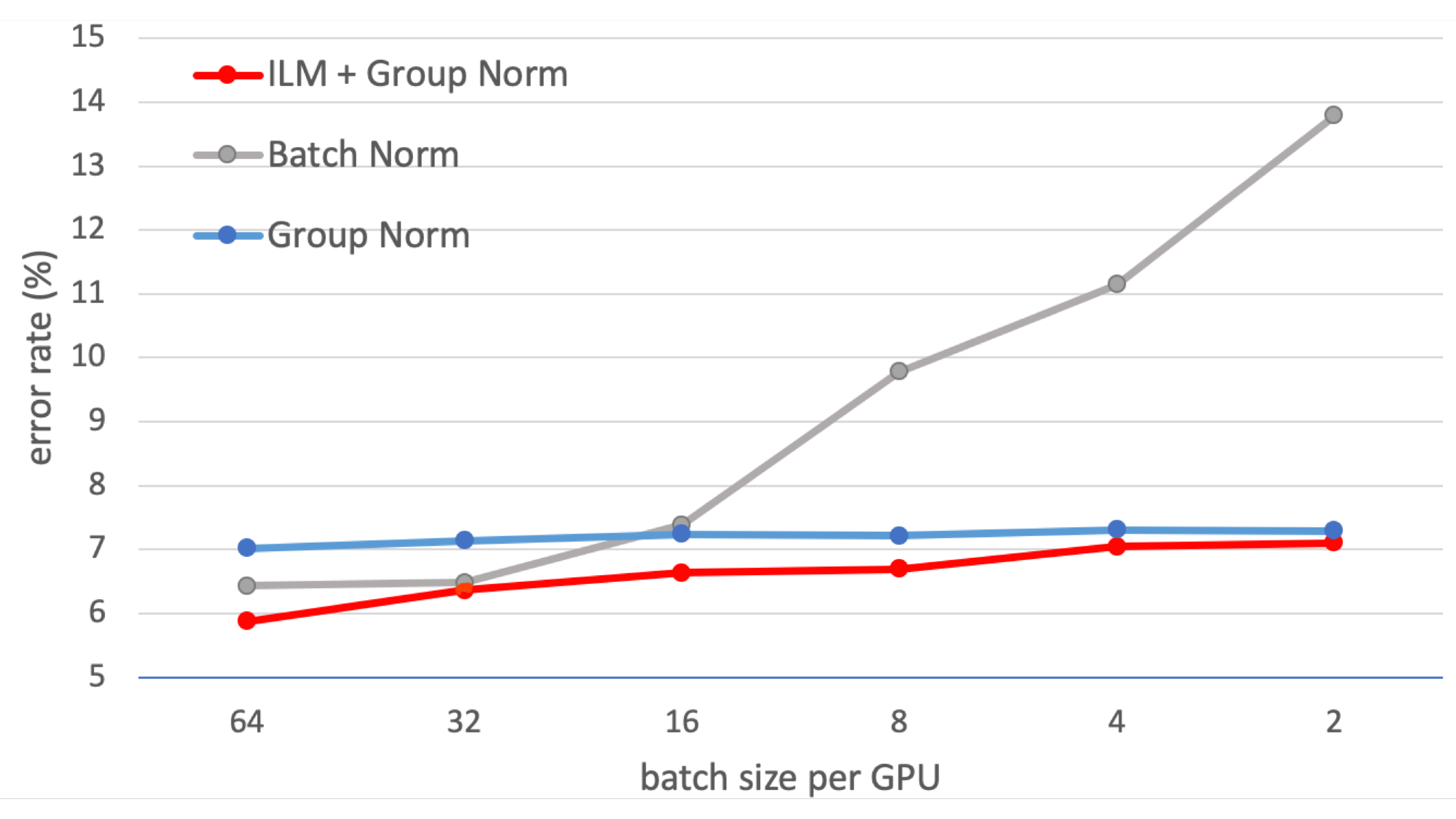}
  \caption{\textbf{CIFAR-10 classification error rate versus batch size per GPU.} The evaluation model is ResNet-101. The result shows that ILM~Norm applied on Group Normalization has the best error rates in comparison with Batch Normalization~\cite{IoffeS15} and the original Group Normalization~\cite{WuH18}.}
  \label{fig:varBatchSize}
\end{figure}

Existing normalization techniques mainly focus on studying the standardization stage to improve the training of deep networks under various circumstances. In contrast, as far as we know, the rescaling stage is less investigated and its related improvements remain unexplored. We observe that existing techniques of estimating the rescaling parameters for recovering the standardized input feature tensor often merely rely on the back-propagation process without considering the correlation between the standardization stage and the rescaling stage.
As a result, information might be lost while data flow is passing through these two stages. We argue that the lack of correlation between two stages may lead to a performance bottleneck for existing normalization techniques.  
 

\begin{figure*}[th]
    \centering
    \includegraphics[width=0.85\textwidth]{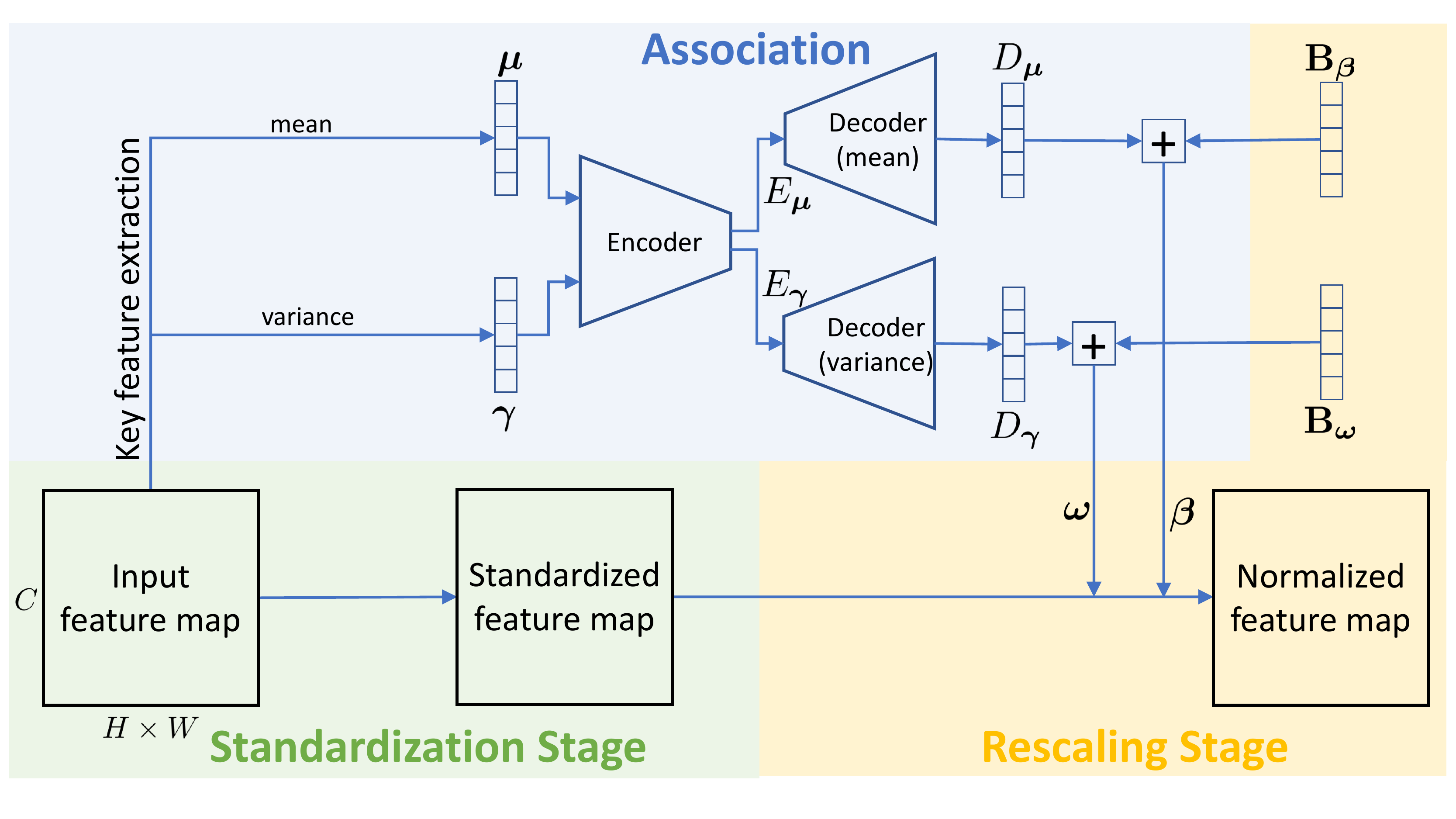}
    
    \caption{An overview of ILM~Norm, the proposed meta learning mechanism for instance-level normalization. As a meta normalization mechanism, ILM~Norm can derive its mechanism for the standardization stage and parameters for the rescaling stage from all kinds of instance-level standardization methods. For association, ILM~Norm divides the $C$ channels of the input into groups for computing the mean and variance per group as the key features $\bmu$ and $\bgamma$. An auto-encoder is then used to associate the features $\bmu$ and $\bgamma$ of the input feature tensor to the rescaling parameters $\bomega$ and $\bbeta$ via the outputs ($D_{\bgamma}$ and $D_{\bmu}$) and the original rescaling parameters ($B_{\bomega}$ and $B_{\bbeta})$ for rescaling the standardized feature map.}
    \label{fig:overview}
\end{figure*}

The proposed Instance-Level Meta Normalization (ILM~Norm) aims to connect the standardization stage and the rescaling stage. The design of ILM~Norm is inspired by residual networks \cite{HeZRS16}, which use the previously visited feature maps for guiding the learning of the current feature maps. The learning of weight $\bomega$ and bias $\bbeta$ in ILM~Norm follows the clue from the input feature maps of the standardization stage rather than merely relying on the back-propagation as previous methods do. We link the input tensor $\rvx$ of feature maps with the weight $\bomega$ and the bias $\bbeta$ in the rescaling stage. In this way, the weight $\bomega$ and the bias $\bbeta$ can not only be optimized better but fit to different inputs more effectively during the forward pass as well. 

ILM~Norm provides a meta learning mechanism for instance-level normalization techniques. It is handy for combining with existing instance-level techniques such as Instance Normalization \cite{UlyanovVL16} or Group Normalization \cite{WuH18}. 
ILM~Norm normalizes the features within each instance individually, \ie, performing the normalization without using the batch dimension. 
The advantage of this property is that, since the normalization is independent of the batch size, the performance is more robust to various settings of batch sizes that are suitable for particular network architectures on different tasks. 


An overview of the proposed meta normalization mechanism is shown in Figure~\ref{fig:overview}. The main ideas, advantages, and contributions of this work are summarized as follows:
\begin{enumerate}
\item ILM~Norm provides a novel way to associate the rescaling parameters with the input feature maps rather than deciding the parameters merely from back-propagation. 
\item ILM~Norm can be handily plugged into existing instance-level normalization techniques such as Instance Normalization \cite{UlyanovVL16}, Layer Normalization \cite{BaKH16}, and Group Normalization \cite{WuH18}. We show that ILM~Norm improves existing instance-level normalization techniques on various tasks.
\item The number of variables in ILM~Norm is small and does not add too much computation burden. For ResNet-101, the total number of variables would only increase 0.086\% with ILM~Norm.
\item The experimental results show that ILM~Norm performs stably well under various batch sizes.
\item We conduct extensive experiments on several datasets to analyze and compare the properties of ILM~Norm with other normalization techniques.
\end{enumerate}



\section{Related Work}

\subsection{\label{sec:NormalizationWork}Normalization in Deep Neural Networks}
Gradient-based learning may suffer from the well-known problems of exploding gradient or vanishing gradient. It has been demonstrated that normalization provides an effective way to mitigate such problems.


Several popular normalization techniques have been proposed with the developments of deep neural networks. AlexNet~\cite{KrizhevskySH12} and its follow-up models \cite{SermanetEZMFL13,SzegedyLJSRAEVR15} adopt Local Response Normalization (LRN) \cite{JarrettKRL09,LyuS08} to compute the mean and variance of the same spatial locations across several neighboring feature maps for standardizing to the middle one. However, this kind of normalization only focuses on the statistics in a small neighborhood per pixel.

As suggested by its name, Batch Normalization (BN) \cite{IoffeS15} provides a batch-level normalization method that respectively centers and scales by mean and variance across the whole mini-batch and then rescales the result. Decorrelated Batch Normalization \cite{HuangYLD18} improves Batch Normalization by adding an extra whitening procedure at the standardization stage. For batch-level normalization mechanisms, the calculation of mean and variance relies on the whole mini-batch. The effectiveness of normalization may degrade when the batch size is not sufficient to support the statistics calculation. To ease the issue of degradation, Batch Renormalization \cite{Ioffe17} suggests adding more learnable parameters in BN.


Several normalization techniques  \cite{ArpitZKG16,BaKH16,UlyanovVL16,RenLUSZ17,WuH18} inherit the notion of Batch Normalization but mainly focus on the manipulations of the standardization stage. Layer Normalization (LN) \cite{BaKH16} operates along the channel dimension and standardizes the features from a single mini-batch by the mini-batch's own mean and variance. It can be used with batch size 1. Instance Normalization (IN) \cite{UlyanovVL16} standardizes each feature map with respect to each sample. Group Normalization (GN) \cite{WuH18} divides the feature channels within each mini-batch into several groups and then performs the standardization for each group. GN's computation is also independent of batch sizes, and we consider it an instance-level normalization technique that can be augmented with ILM~Norm.

Another way to do normalization is adjusting the filter weights instead of modifying the feature maps. For example, Weight Normalization \cite{SalimansK16} and Orthogonal Weight Normalization \cite{HuangLLYWL18} present this kind of normalization strategy to address some recognition tasks.


We observe that the existing normalization methods merely focus on manipulating the learning of parameters at the \emph{standardization stage}. They do not consider the correlations between the standardization stage and the rescaling stage. The parameters learned for rescaling are based on back-propagation and might be of low correlation with the parameters for standardization. Our experimental results show that taking into account the connection between standardization and rescaling is beneficial.

\subsection{Style Transfer with Rescaling Parameters}

The goal of a style transfer task is to `extract' or `imitate' a visual style from one image and apply that style to another image. Likewise, domain adaptation aims to enable a function learned from one domain to work comparably well in another domain. One solution to this kind of task is manipulating the learned rescaling parameters, and therefore we quickly review some style transfer methods that are related to learning rescaling parameters. 

The core idea of using the learned rescaling parameters to address the tasks of style transfer or domain adaptation is similar to the normalization process. The original distribution of one domain is standardized and then mapped to the target distribution in the target domain. Hence, the rescaling parameters learned from the target distribution can be used to recover the original distribution in the target domain.

Adaptive Instance Normalization \cite{HuangB17} applies the rescaling parameters generated by another domain to the feature maps of the current domain via Instance Normalization. Dynamic Layer Normalization \cite{KimSB17} generates the rescaling parameters by different speakers and environments for adaptive neural acoustic modeling via Layer Normalization. 


\section{Instance-Level Meta Normalization}


This section describes the proposed two-stage learning mechanism for improving instance-level normalization. Our approach is applicable to various techniques that perform instance-level normalization, and hence we call it \emph{Instance-Level Metal Normalization} (ILM~Norm). Figure~\ref{fig:overview} shows an overview of ILM~Norm. The first stage is \emph{standardization}, which regularizes the mean $\bmu$ and variance $\bgamma$ of the input feature tensor $\rvx$ for standardizing the distribution of the feature tensor. The second stage is \emph{rescaling}, which rescales the standardized feature map $\rvx_s$ for recovering the representation capability of the feature tensor $\rvx$. Moreover, we employ an auto-encoder to serve as an \emph{association} between two stages. The rescaling stage uses the auto-encoder to predict the rescaling parameters, \ie, weight $\bomega$ and bias $\bbeta$, with respect to the input tensor $\rvx$ of feature maps instead of generating the rescaling parameter simply from back-propagation.

\subsection{Standardization Stage}
The goal of the standardization stage is to regularize the distribution of the input feature map, which is often done by forcing the distribution to have zero mean and unit variance. Existing normalization techniques mostly focus on designing different schemes for this stage.

As a meta learning mechanism, ILM~Norm can adopt different standardization processes from different instance-level normalization techniques. Take, for example, Group Normalization's standardization process. 
Group Normalization (GN) divides the whole layer into several groups along its channel dimension. Each group calculates its own mean and variance for standardization. 
Many other methods can be considered. It is free to be replaced by others for different purposes.

\subsection{Rescaling Stage}
The goal of the rescaling stage is to recover the distribution of the input feature maps from its standardized counterpart. Previous approaches usually learn the parameters for recovering the statistics merely via back-propagation. In contrast, ILM~Norm predicts the parameters with additional association between the standardization stage and the rescaling stage. In the following, we detail the process of extracting the key features of the input tensor $\rvx$ of feature maps. The auto-encoder for predicting the rescaling parameters will be presented in Section~\ref{sec:autoencoder}.


\subsection{\label{sec:autoencoder}Association between Two Stages}
The association between the standardization stage and the rescaling stage is achieved by a coupled \emph{auto-encoder}, which is of little computational cost.
An overview of the components is shown in Figure~\ref{fig:overview}. ILM~Norm contains an auto-encoder for predicting the rescaling parameters $\bomega$ and $\bbeta$ concerning the pre-computed mean $\bmu$ and the variance $\bgamma$ of the input feature tensor $\rvx$. In comparison with existing methods that simply learn the parameters $\bomega$ and $\bbeta$ via back-propagation, our experiments show that the meta parameter learning mechanism with additional information from $\bomega$ and $\bbeta$ is more effective. 

\subsubsection{\label{sec:keyfeature}Key Feature Extraction}

ILM~Norm uses an auto-encoder to predict the weights $\bomega$ and bias $\bbeta$ as the rescaling parameters for recovering the distribution of the tensor $\rvx$ of feature maps. We have observed that directly encoding the entire input feature tensor $\rvx$ would degrade the prediction accuracy, which might be due to overfitting. Instead of using the entire feature tensor $\rvx$ as the input for the auto-encoder, we propose to use the mean $\bmu$ and variance $\bgamma$ of $\rvx$ for characterizing its statistics. Here we define the key features as the mean $\bmu$ and variance $\bgamma$ extracted from the feature tensor $\rvx$. The experimental results demonstrate that, ILM~Norm, which uses a light-weight auto-encoder, can effectively predict $\bomega$ and $\bbeta$ for recovering the distribution of input tensor $\rvx$ of feature maps.

Furthermore, for better performance and lower computation burden, we extract the key features from each group of input feature maps rather than a single feature map. For a specific layer comprising $C$ channels as a tensor of feature maps $f_1, f_2, \dots, f_C$, we evenly partition these feature maps into $N$ groups $\mathbf{f}_1, \mathbf{f}_2, \dots, \mathbf{f}_N$. The mean and variance of the whole layer are hence denoted as a vector of length $N$, namely $\bmu = [\mu_1, \mu_2, \dots, \mu_N]$ and $\bgamma = [\gamma_1, \gamma_2, \dots, \gamma_N]$. ILM~Norm computes the mean $\mu_n$ and variance $\gamma_n$ for a given feature-map group $\mathbf{f}_n$ as 
\begin{equation}
    \left\{ \begin{array}{l@{\;}l}
        \mbox{${\mu}_n = \frac{1}{H \times W \times C/N}\sum_{f \in \mathbf{f}_n}\sum_{i=1}^{H}\sum_{j=1}^{W} f^{i,j}$}\,, \\
        \mbox{$\gamma_n = \frac{1}{H \times W \times C/N}\sum_{f \in \mathbf{f}_n}\sum_{i=1}^{H}\sum_{j=1}^{W} (f^{i,j}-\mu_n)^2$}\,,
    \end{array} \right.
\end{equation} 
\noindent where $C/N$ is the number of feature maps in a group, $f$ denotes a feature map of group $\mathbf{f}_n$. Further discussions for key feature extraction can be found in Section~\ref{sec:keyFeature}.

\comment{
\begin{equation}
\${\mu}_k = \mathbf{F}_{\mathit{mean}}(\mathbf{u}_k) = \frac{g}{H \times W \times C}\sum_{G_n=G_k}\sum_{i=1}^{H}\sum_{j=1}^{W}(\mathbf{u}_{n}^{i,j})
\end{equation}
\begin{equation}
\{\gamma_k = \mathbf{F}_{\mathit{var}}(\mathbf{u}_k) = \frac{g}{H \times W \times C}\sum_{G_n=G_k}\sum_{i=1}^{H}\sum_{j=1}^{W}(\mathbf{u}_{n}^{i,j} - \boldsymbol{\mu}_k)^2
\end{equation}
}


\subsubsection{Encoder} 

The goal of the encoder in ILM~Norm is to summarize the information of an input tensor‘s key features through an embedding. Besides, we expect the subsequent rescaling parameters can be jointly learned from the same embedded information.  

In our implementation, the encoder comprises one \emph{fully connected layer} ($\mathbf{W}_1$) and one \emph{activation function}. The fully connected layer can model not only the individual elements of the key features but also the correlations between elements. Using an activation function allows us to extract non-linear information. The embedded vectors, which encode the mean and the variance of the grouped input feature maps, are obtained by 
\begin{equation}
    \left\{ \begin{array}{l@{\;}l}
        \mbox{$E_{\bmu} = \mathrm{ReLU}(\mathbf{W}_{1} \bmu)$}\,, \\
        \mbox{$E_{\bgamma} = \mathrm{ReLU}(\mathbf{W}_{1} \bgamma)$}\,,
    \end{array} \right.
\end{equation} 
\noindent where $E_{\bmu}$ and $E_{\bgamma}$ respectively denote the embedded vectors of $\bmu$ and $\bgamma$, $\mathrm{ReLU}(\cdot)$ represents the activation function, and the encoding matrix $\mathbf{W}_{1} \in {\mathbb{R}}^{M \times N}$ with the embedded vector of length $M$ and key feature vectors of length $N$.

\subsubsection{Decoder} 

\comment{What the decoder does is as follows:
\begin{equation}
D_{\boldsymbol{\mu}} = \delta_{sigmoid}(\textbf{W}_{2} \textbf{E}_{\boldsymbol{\mu}})
\end{equation}
\begin{equation}
D_{\boldsymbol{\gamma}} = \delta_{Tanh}(\textbf{W}_{3} \textbf{E}_{\boldsymbol{\gamma}})
\end{equation}
\noindent where $D_{\boldsymbol{\mu}}$ and $D_{\boldsymbol{\gamma}}$ are the vectors decoded from $E_{\boldsymbol{\mu}}$ and $E_{\boldsymbol{\gamma}}$, and $\textbf{W}_{2}$, $\textbf{W}_{3} \in {\mathbb{R}}^{g,e}$.}

\comment{\djc{The decoder in AssocNorm aims to predict the rescaling parameters $\boldsymbol{\omega}$ and $\boldsymbol{\beta}$ from the embedded vectors $E_{\boldsymbol{\mu}}$ and $E_{\boldsymbol{\gamma}}$, respectively. }}

The decoder in ILM~Norm aims to decode the embedded vectors $E_{\bmu}$ and $E_{\gamma}$ into $D_{\bmu}$ and $D_{\bgamma}$ respectively. In a sense, $D_{\bmu}$ and $D_{\bgamma}$ propagate the correlations from the original feature maps to the rescaling parameters $\bomega$ and $\bbeta$.

In our implementation, we use two different \emph{fully connected layers} ($\mathbf{W}_2$ and $\mathbf{W}_3$) and two \emph{activation functions}. The fully connected layers for decoding aim to summarize the information-rich embedded vectors for predicting the rescaling parameters. By accompanying the decoded vector with an activation function, ILM~Norm shifts the vector values into a suitable range. The decoded vectors, which yield the mean and variance of the embedded vector, are obtained as
\begin{equation}
    \left\{ \begin{array}{l@{\;}l}
        \mbox{$D_{\bmu} = \tanh(\mathbf{W}_{2} E_{\bmu})$}\,, \\
        \mbox{$D_{\bgamma} = \mathrm{sigmoid}(\mathbf{W}_{3} E_{\bgamma})$}\,,
    \end{array} \right.
\end{equation} 
\noindent where both $\mathrm{sigmoid}(\cdot)$ and $\tanh(\cdot)$ represent the activation functions, and the decoding matrices $\mathbf{W}_{2}$, $\mathbf{W}_{3} \in {\mathbb{R}}^{N \times M}$. Further discussions about choosing activation functions can be found in Section~\ref{sec:activationfunctions}.



\subsubsection{Alignment\label{sec:bias}}
Notice that each decoded vector of $D_\bmu$ and $D_\bgamma$ predicted from the auto-decoder needs to align with a corresponding rescaling parameters of the underlying normalization module in which the ILM~Norm is plugged, , \eg, Instance Normalization. We obtain the final rescaling parameters $\bomega$ and $\bbeta$ as follow:
\begin{equation}
    \left\{ \begin{array}{l@{\;}l}
        \mbox{$\bomega = D_\bgamma\uparrow + B_\bomega$}\,, \\
        \mbox{$\bbeta = D_\bmu\uparrow + B_\bbeta$}\,,
    \end{array} \right.
\end{equation}      
\noindent where $B_\bomega$ and $B_\bbeta$ denote the rescaling parameters of the underlying normalization module that is augmented by ILM~Norm. The dimension of either $B_\bomega$ or $B_\bbeta$ is $C$, \ie, the number of channels. The operator $\uparrow$ means duplicating the vector components so that the dimension of $D_\bgamma$ and $B_\bomega$ can match, and same for $D_\bmu$ and $B_\bbeta$.


\begin{figure*}[t]
    \centering
    \includegraphics[width=\textwidth]{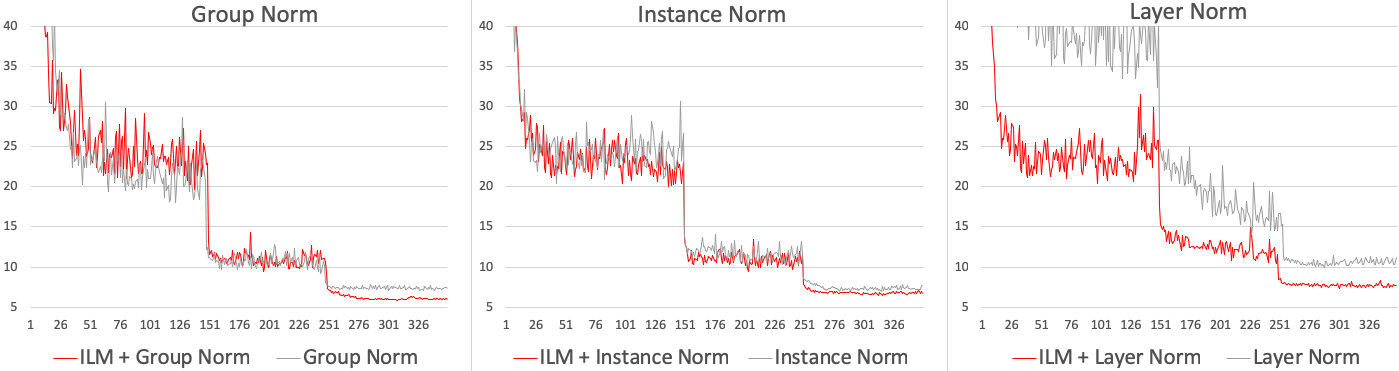}
    \caption{Comparisons of different instance-level normalization techniques. We show the validation error rate (\%) against the number of training epochs. The batch size is $64$. The performances of the original normalization techniques are improved after applying ILM~Norm.}
    \label{fig:largeBatchSize}
\end{figure*}

\section{Experiments}
In the experiments we evaluate ILM~Norm using different datasets on various tasks. We apply ILM~Norm to several state-of-the-art instance-level normalization techniques, including Layer Normalization (LN), Instance Normalization (IN), and Group Normalization (GN), and we show that the ILM~Norm enhanced versions steadily outperform the original ones.

\subsection{Image classification with Large Batch Size}

\subsubsection{Implementation Details}

We use ResNet-50 and ResNet-101 \cite{HeZRS16} as a backbone model for evaluating the experiments on classification tasks. For CIFAR-10 and CIFAR-100 datasets, we change the first conv layer to ``$3 \times 3$, stride 1, padding 1'', remove the max-pooling layer, and change the kernel size of average-pooling to 4 to adapt to the input size. We initialize all parameters using the standard normal distribution except the rescaling parameters of the corresponding underlying normalization module $B_\bomega$ and $B_\bbeta$, which are assigned as $1$ and $0$, respectively.

Unless otherwise stated, ILM~Norm set the size of group equals to $16$ (\ie, $C/N=16$), the number $N$ of groups of GN is set to $32$, and the batch size of all normalization methods is $ 64$. We use SGD as the optimizer with momentum $0.9$ and weight decay $0.0005$. All experiment are conducted on only one GPU. For CIFAR-10 and CIFAR-100 datasets, each normalization method is trained for 350 epochs. The learning rate is initialized with $0.1$ and decreased by $0.1$ at the $150$th and $250$th epoch. For ImageNet, each normalization method is trained for 100 epochs. We set learning rate as $0.025$ according to the suggestion of \cite{GoyalDGNWKTJH17}. The learning rate is decreased by $0.1$ at $30$th, $60$th and $90$th epoch.

\begin{table}[t]
   \centering
    \begin{tabular}{|p{2.5cm}|c|c|c|c|}
    \hline
    \multirow{2}{*}{Top-1 Error ($\%$)} & \multicolumn{4}{c|}{Method} \\ \cline{2-5} 
              & BN   & GN   & IN   & LN   \\ \hline
     Original    & $6.43$  & $7.02$ & $7.00$ & $9.98$ \\ \hline
     with ILM   & - & $5.88$  & $6.50$  & $7.35$ \\ \hline \hline
     with ILM vs. Original   & - & $-1.14$ & $-0.50$ & $-2.63$ \\ \hline
    with ILM vs. BN &    -  & $-0.55$ & $+0.07$ & $+0.92$ \\ \hline
    \end{tabular}
    \caption{Comparison of different normalization methods on CIFAR-10. }
    \label{tab:largeBatchSize}
\end{table}


\subsubsection{CIFAR-10}
 We compare several instance-level normalization methods (GN, IN, LN) with their ILM~Norm extensions for image classification on CIFAR-10 dataset \cite{Krizhevsky09,TorralbaFF08}. The underlying architecture is ResNet-101. We also present the result of Batch Normalization (BN) trained under the same configuration as a strong baseline. The results are shown in Figure~\ref{fig:largeBatchSize} and Table~\ref{tab:largeBatchSize}.

Figure~\ref{fig:largeBatchSize} shows the comparisons of different instance-level normalization techniques. We plot the validation error rate against the number of training epochs. \comment{The batch size is $64$.} ILM~Norm is applied to IN, GN and LN, and all the three normalization methods can be improved to achieve lower validation error rates.
 
Table~\ref{tab:largeBatchSize} shows the error rates of different methods with $350$ training epochs. Notice that, in the last two rows of Table~\ref{tab:largeBatchSize} we compare the change in performance after applying ILM~Norm. We show the relative error rate w.r.t. the original normalization and w.r.t. BN. As can be seen, all instance-level normalization methods can achieve a lower error rate after being equipped with ILM~Norm. Furthermore, the combination of ILM+GN can even outperform BN. It is worth mentioning that, to our best knowledge, no existing state-of-the-art instance-level normalization methods have outperformed BN when a large batch size is used on the CIFAR-10 classification task.

\begin{table}[t]
    \centering
    \begin{tabular}{|c|c|c|c|c|}
    \hline
    \multirow{2}{*}{Top-1 Error $(\%)$} & \multicolumn{4}{c|}{Method} \\ \cline{2-5} 
              & BN    & GN    & IN    & LN   \\ \hline
     Original    & $26.28$ & $26.94$ & $26.08$ & $41.61$ \\ \hline
     + ILM        & -     & $23.31$ & $23.97$ & $25.43$ \\ \hline \hline
     +ILM vs. Original & -     & $-3.63$ & $-2.11$ & $-16.18$\\ \hline
     +ILM vs. BN &    -     & $-2.97$ & $-2.31$ & $-0.85$ \\ \hline
    \hline
    \hline
    \multirow{2}{*}{Top-5 Error $(\%)$} & \multicolumn{4}{c|}{Method} \\ \cline{2-5} 
             & BN    & GN    & IN    & LN   \\ \hline
     Original    & $9.37$ & $7.02$  & $7.60$  & $15.26$ \\ \hline
      +ILM        & -  & $6.47$ & $6.71$ & $6.88$ \\ \hline \hline
     +ILM vs. Original & - & $-0.55$ & $-0.89$ & $-8.38$\\ \hline
     +ILM vs. BN &    -     & $-2.90$ & $-2.66$ & $-2.49$ \\ \hline
    \end{tabular}  
    \caption{Comparison of different normalization methods on CIFAR-100. 
    \label{tab:comCifar100}}
\end{table}

\subsubsection{CIFAR-100}

 We conduct another 
 similar experiment to compare different normalization methods on CIFAR-100~\cite{Krizhevsky09,TorralbaFF08} image classification task. All models are trained on the training set of $50{,}000$ images and evaluated on the validation set of $10{,}000$ images for 350 epochs. The results are shown in Table~\ref{tab:comCifar100}. Similar improvements on the CIFAR-100 classification task for different methods can be observed.

\begin{table}[t]  
    \centering
    \begin{tabular}{|c|c|c|c|c|c|}
    \hline
    \multirow{2}{*}{ImageNet} & \multicolumn{5}{c|}{Method (Error Rate $\%$)} \\ \cline{2-6} 
              & BN & ILM+GN & GN   & IN   & LN   \\ \hline
     Top-1    & $23.85$  & $23.57$ & $24.06$ & $28.40$ & $25.30$ \\ \hline \hline
    vs. BN &    -  &$-0.28$ & $+0.21$ & $+4.55$ & $+1.45$ \\ \hline
    \end{tabular}
    \caption{Comparison of different normalization methods on ImageNet. }
    \label{tab:ImageNet}
\end{table}

\subsubsection{ImageNet}
We also use ImageNet to evaluate the setting of ILM~Norm plus GN (ILM+GN), in comparison with other normalization methods including BN, IN, LN, and the original GN. The underlying network architecture is ResNet-50. The ImageNet dataset contains over one million images with 1000 different classes. All of the models are trained on the ImageNet training set and evaluated on the validation set. The results are in Table~\ref{tab:ImageNet}.

Table~\ref{tab:ImageNet} shows the error rates after 100 training epochs for different normalization methods. We can find that ILM+GN achieves a $0.49\%$ lower error rate than the original GN. Moreover, ILM+GN achieves a $0.28\%$ lower error rate than Batch Normalization as well, while the basic instance-level normalization methods cannot outperform BN on this task.

In sum, the experiments on classification tasks with CIFAR-10, CIFAR-100, and ImageNet demonstrate that instance-level normalization methods, such as GN, IN, and LN, can be improved if they are equipped with ILM Norm. Furthermore, ILM+GN is able to achieve better performance than cross-instance normalization like Batch Normalization for a large-batch-size setting on ImageNet, which has never been reported before according to our best knowledge. The advantage of ILM Norm for various instance-level normalization methods is therefore evident, and the improvement can be handily achieved with a negligible computation overhead.

\subsection{Image Classification with Various Batch Sizes}
The batch size is an issue to be taken into consideration when apply normalization techniques. 
We conduct an experiment to evaluate ILM Norm plus GN for various batch sizes on CIFAR-10. 
We test the batch sizes of $\{64, 32, 16, 8, 4, 2 \}$ per GPU, without changing other hyper-parameters. For comparison, we also include the results of BN. The error rates are shown in Table~\ref{tab:varBatchSize} and Figure~\ref{fig:varBatchSize}.

Figure~\ref{fig:varBatchSize} clearly illustrates that both GN and ILM+GN are not sensitive to the batch size. Furthermore, ILM+GN gets lower validation error rates than GN among all kinds of batch sizes. In contrast, BN obviously requires a larger batch size and gets considerable large error rates when the batch size is small.

Table~\ref{tab:varBatchSize} shows that ILM+GN has the lowest error rates among all batch sizes. On average, ILM plus GN achieves a lower error rate than GN by $0.58\%$ and also a lower error rate than BN by $2.55\%$ among the evaluated batch sizes.  

\begin{table*}[t]
    \centering
    \begin{tabular}{|c|c|c|c|c|c|c|c|}
    \hline
    \multicolumn{2}{|c|}{\multirow{2}{*}{CIFAR-10: Error Rate $(\%)$} } & \multicolumn{6}{c|}{Batch Size} \\ \cline{3-8} 
    \multicolumn{2}{|c|}{}                  & 64  & 32  & 16  & 8  & 4  & 2  \\ \hline  
    \multirow{3}{*}{Method}     &   GN    &  $7.02$ &  $7.14$ &  $7.24$ &  $7.22$ &  $7.31$ &  $7.29$  \\ \cline{2-8} 
                                &   BN    &  $6.43$ & $6.48$ &  $7.39$ &  $9.78$ & $11.15$ & $13.79$  \\ \cline{2-8} 
                                 & ILM+GN  &$\mathbf{5.88}$ &  $\mathbf{6.36}$ &$\mathbf{6.64}$ &$\mathbf{6.70}$ &$\mathbf{7.05}$ &$\mathbf{7.11}$ \\ \cline{2-8} 
                                \hline
                                \hline
    \multirow{2}{*}{Improvement}          
                                &  ILM+GN vs. GN & $-1.14$ & $-0.78$ & $-0.60$ & $-0.52$ & $-0.26$ & $-0.18$  \\ \cline{2-8} 
                                & ILM+GN vs. BN & $-0.55$ & $-0.12$ & $-0.75$ & $-3.08$ & $-4.10$ & $-6.68$  \\ \hline
    \end{tabular}
    \caption{Evaluations on CIFAR-10 dataset with different batch sizes.  
    \label{tab:varBatchSize}}
\end{table*}

\begin{table*}[tb]
    \centering
    \begin{tabular}{|c|c|||c|c|c||c|c|c|}
    \hline
                Batch Size & Box Head  & $AP^{bbox}$ & $AP^{bbox}_{50}$ & $AP^{bbox}_{75}$ & $AP^{mask}$ & $AP^{mask}_{50}$ & $AP^{mask}_{75}$ \\ \hline \hline
      \multirow{2}{*}{2} & GN    & $22.40$ & $37.65$&                           $23.56$ & $20.88$   &  $35.43$ &                      $21.57$   \\ \cline{2-8}
                          & ILM + GN & $\mathbf{22.68}$ & $\mathbf{38.28}$ & $\mathbf{23.68}$ & $\mathbf{21.19}$&  $\mathbf{36.08}$ & $\mathbf{21.92}$   \\  \cline{2-8}\hline \hline
      \multirow{2}{*}{16} &GN    & $39.10$ & $60.33$&                          $42.51$ & $34.77$   &  $56.88$ &                      $36.79$   \\ \cline{2-8}
                         & ILM + GN & $\mathbf{39.42}$ & $\mathbf{60.63}$ & $\mathbf{42.95}$ & $\mathbf{35.03}$&  $\mathbf{57.25}$ & $\mathbf{36.92}$   \\ \hline
    \end{tabular} 
    \caption{Evaluations on MS-COCO dataset for detection and segmentation tasks with different batch sizes. 
    \label{tab:Maskrcnn}}   
\end{table*}

\begin{table}[tb] 
    \centering
    \begin{tabular}{|c|c|c|c|}
    \hline
    \multirow{2}{*}{ 
    Metric} & \multicolumn{3}{c|}{Method (Generator / Discriminator)} \\ \cline{2-4} 
              & IN / IN   & ILM+IN / IN   & ILM+IN / ILM+IN   \\ \hline
     RMSE    & $108.17$  & $105.82$ &  $\mathbf{105.46}$ \\ \hline
     LPIPS    & $0.441$  & $0.435$ & $\mathbf{0.428}$ \\ \hline \hline
    SSIM    & $0.372$  & $\mathbf{0.390}$ & $0.372$ \\ \hline
    \end{tabular}
   \caption{Evaluation on the Facades dataset for the style transfer task. Note that,  typically, a higher SSIM score means higher similarity, while a lower RMSE or LPIPS value implies better performance.
    \label{tab:imagetransfer}}
\end{table}

\begin{table*}[t]
   \centering
    \begin{tabular}{
    |c|c|c|c|c|c|c|c|}
    \hline
    \multirow{2}{*}{} & \multicolumn{6}{c|}{Activation Functions for $(D_{\bmu},D_{\bgamma})$} & \multirow{2}{*}{GN}  \\ \cline{2-7} 
     &  ours   & (sigmoid, sigmoid)& (tanh, tanh) & ($\dagger$, $\star$) & ($\star$, $\dagger$) & ($\star$, $\star$)  & {}  \\ \hline
     Error Rate $\%$ & $\textbf{5.88}$ & $8.63$ & $8.04$ & n/a & n/a & n/a & 7.02 \\ \hline
    \end{tabular}
    \caption{Comparison of different activation functions for $(D_{\bmu},D_{\bgamma})$ on CIFAR-10. The symbol $\dagger$ means the activation function is either tanh or sigmoid, while $\star$ means the activation function can be ReLU, Leaky ReLU, ReLU6, or Identity. The entry `n/a' indicates that the model cannot converge. }
    \label{tab:diff_acti}
\end{table*}

\begin{table*}[bt]
    \centering
    \begin{tabular}{|c|c|c|c|c|c|}
    \hline
    \multirow{2}{*}{Increment Ratio of Parameters} & \multicolumn{5}{c|}{Model} \\ \cline{2-6} 
                            & ResNet-18 & ResNet-34 & ResNet-50 & ResNet-101 & ResNet-152 \\ \hline
     Group Normalization    & $0.015\%$   & $0.015\%$   & $0.086\%$    & $0.086\%$   &  $0.086\%$   \\ \hline
     Instance Normalization & $2.792\%$   & $2.462\%$    & $20.696\%$    & $20.313\%$   &  $20.178\%$   \\ \hline
    \end{tabular} 
    \caption{The increment in the number of parameters concerning different key-feature extraction strategies. ILM Norm chooses to use GN's scheme for key-feature extraction since the increment in number of additional parameters is less than $0.1\%$ for most of the ResNet models.
    \label{tab:GNIN}}   
\end{table*}

\paragraph{Discussion.}
Table~\ref{tab:varBatchSize} shows that ILM+GN outperforms GN among all batch sizes. Since all hyper-parameters of ILM+GN are set the same as BN, it is reasonable to consider that the improvement is owing to the association mechanism of ILM~Norm that connects the standardization stage and the rescaling stage. As a result, it is helpful to leverage both the cross-stage association and the back-propagation process while learning the rescaling parameters for normalization. 

\comment{
The experiments demonstrate that instance-level normalization methods can achieve better performance on error rate when equipped ILM but not loss batch-size robustness. Even with the large batch size, instance-level normalization methods with ILM can even surpass the BN, which prefers a large batch size, to be the new state-of-the-art in the CIFAR-10 classification task.  }

\subsection{Object Detection and Segmentation}
Object detection and segmentation are important tasks in computer vision. We evaluate ILM~Norm on Mask R-CNN~\cite{HeGDG17} using MS-COCO dataset~\cite{LinMBHPRDZ14}. All models are trained on the training set for $90{,}000$ with batch size per GPU equal to 2 using 1 GPU and 8 GPUs. The backbone used in all models are pretrained with GN. We test the models on the test set. All other configurations are just the same as R-50-FPN in Detectron~\cite{Detectron2018}. The results are shown in Table~\ref{tab:Maskrcnn}.

Table~\ref{tab:Maskrcnn} shows that only changing GN layers in the box head can improve the detection and segmentation performance under different batch sizes. To be more specific, we increase AP$^{bbox}$ and AP$^{mask}$ by 0.28 and 0.31 when the batch size is equal to 2, and by 0.32 and 0.26 when batch size is equal to 16. It indicates that ILM+GN can transfer the feature from backbone more efficiently than GN, compared with the GN baseline, with different training lengths.

\subsection{Image Transfer}
Image transfer is a popular and interesting task in computer vision. We evaluate ILM~Norm on pix2pix \cite{IsolaZZE16} using the CMP Facades \cite{TylecekS13} dataset with size of group equals to 1. The Facades dataset contains 400 architectural-labels-to-photo data. We train the model in 200 of them and evaluate on the rest 200 data for 200 epochs. To evaluate the performance, we use SSIM~\cite{WangBSS04}, RMSE, and the LPIPS metric~\cite{ZhangIESW18} as the similarity measures. Typically, a higher SSIM score means higher similarity, while a lower RMSE or LPIPS value implies better quality. The results are shown in Table~\ref{tab:imagetransfer}.

Table~\ref{tab:imagetransfer} clearly shows that changing all IN layers in the model or only changing IN in the generator can both improve the similarity between the output of the model and the target. Since LPIPS focuses on not only the structural similarity but also the perceptual similarity, using ILM+IN can produce style transfer results with better structural and perceptual quality than original IN.


\subsection{Ablation Study}
\subsubsection{\label{sec:activationfunctions}Different Activation Functions for $D_{\bmu}$ and $D_{\bgamma}$}
Since the rescaling parameters are the only part in the model that is modified during forward propagation, it is critical to control the extent of their variations. Excessive variations in rescaling parameters lead to instability of the model. Moreover, the domain of $\bmu$ and $\bgamma$ are different; it is reasonable to control $D_{\bmu}$ and $D_{\bgamma}$ within a different range. To verify our assumptions, we evaluate ILM Norm with several different activation functions applied on $D_{\bmu}$ and $D_{\bgamma}$. The results can be found in Table~\ref{tab:diff_acti}.

Table~\ref{tab:diff_acti} shows that the model cannot converge without appropriate constraints on $D_{\bmu}$ and $D_{\bgamma}$. Applying the same activate function, \eg sigmoid, to both $D_{\bmu}$ and $D_{\bgamma}$ may make the model converge, but the performance is even worse than the original group normalization, indicating that the association has a negative impact on the normalization. Only by deploying different activation functions, tanh to $D_{\bmu}$ and sigmoid $D_{\bgamma}$ can we achieve positive impact and the best performance among these configurations.

\subsubsection{\label{sec:keyFeature}Alternative Strategies for Key-Feature Extraction}
\begin{figure}[b]
    \centering
    \includegraphics[width=1.1\linewidth]{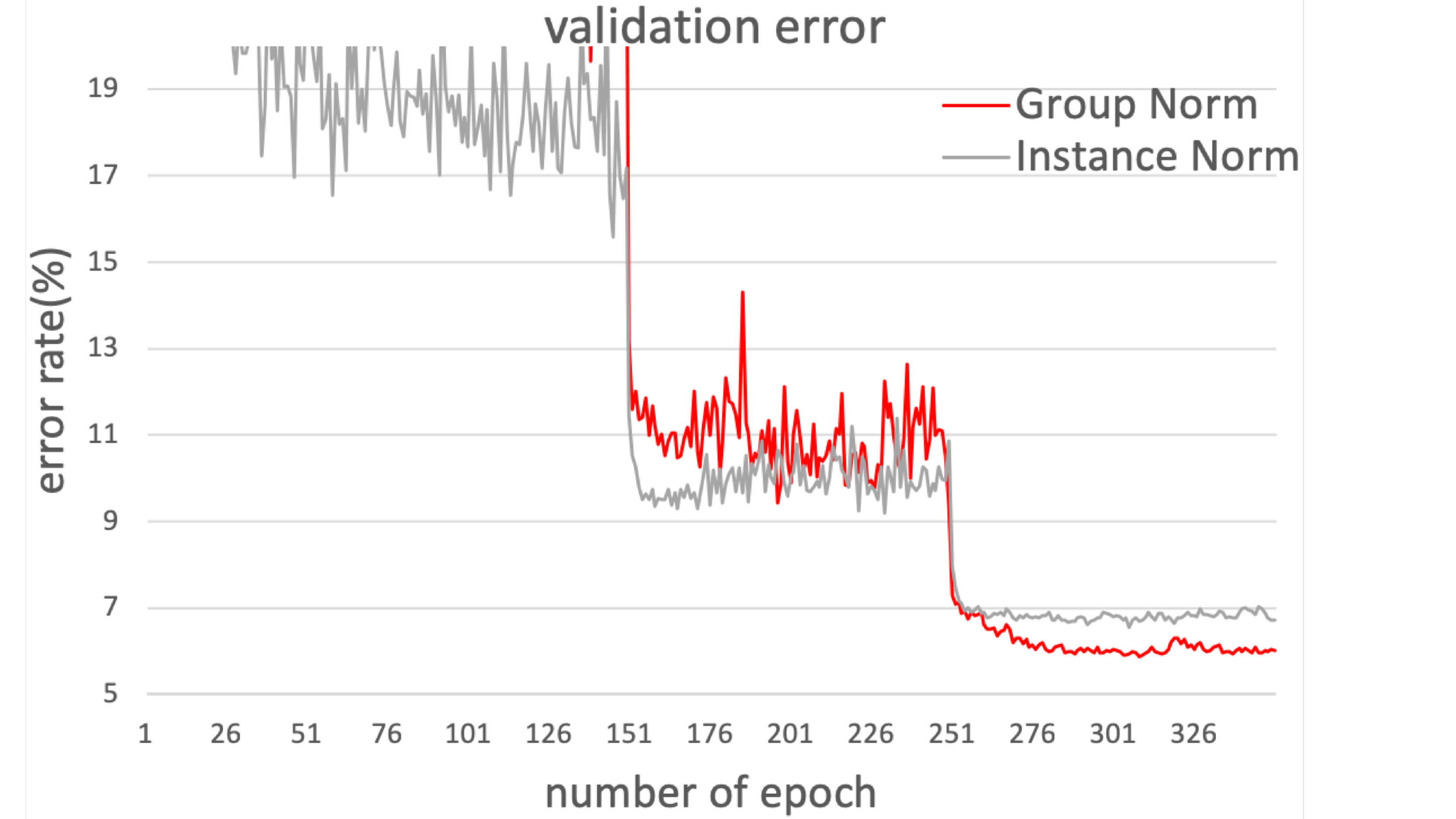}
    \caption{Comparison on the performance of using a key-feature extraction strategy as GN or IN. The evaluation is based on CIFAR-10 validation error.}
    \label{fig:GNIN}
\end{figure}

ILM Norm divides the input channels into groups for computing the mean and variance per group. As mentioned in Section~\ref{sec:NormalizationWork}, the existing normalization methods, such as BN, LN, IN, and GN have their own scheme to extract the mean $\bmu$ and variance $\bgamma$ from the input $\rvx$. To make our normalization mechanism robust to various batch sizes, we do not consider the scheme of BN. Moreover, we also exclude LN's scheme, since LN only generates one pair of mean and variance for all feature maps in a layer, and such few data is not suitable for training our auto-encoder-like network. Therefore, our key-feature extraction strategy considers the implementations as IN and GN.
Figure~\ref{fig:GNIN} shows a comparison of the key feature extraction strategies derived from GN and IN. Table~\ref{tab:GNIN} provides the ratio of increment in the number of parameters concerning different key-feature extraction strategies.

Figure~\ref{fig:GNIN} shows that the performance of using a key-feature extraction strategy as GN is usually better than the performance of using IN. From the perspective of the increment in the number of parameters, Table~\ref{tab:GNIN} provides further information for choosing the strategy of key feature extraction. In Table~\ref{tab:GNIN}, the number of additional parameters due to the use of the key-feature extraction strategy as GN is quite small. The lower requirement of additional parameters for GN is because it partitions the $C$ channels into $N$ groups, where $N = C/K$ and the size of a group is fixed to $K$, and hence the increment in the number of parameters depends on the ratio $C/K$ instead of $C$. In contrast, the increment in the number parameters using IN's scheme depends on $C$. 

To sum up: The experiments demonstrate that using the key-feature extraction strategy as GN should be the best option. It not only achieves a lower error rate but also requires less increment in the number of parameters.

\section{Conclusion}

We have presented ILM Norm, a meta learning mechanism for various instance-level normalization techniques. ILM Norm extracts the key features from the input tensor and associates the standardization parameters with the rescaling parameters for deep network normalization. As a result, ILM Norm provides an easy way to predict the rescaling parameters via both the update from back-propagation and the association with input features. ILM works well with state-of-the-art instance-level normalization methods, and meanwhile, improves the performance in most cases. The experiments demonstrate that a deep network equipped with ILM Norm is able to achieve better performance for different batch sizes with just a little increase in the number of parameters.


\paragraph{Acknowledgement:}
This work was supported in part by MOST Grants 108-2634-F-001-007 and 106-2221-E-007-080-MY3.
We thank Ting-I Hsieh, Yi-Chuan Chang, Wen-Chi Chin, and Yi-Chun Lin for insightful discussions.
We are also grateful to the {\em National Center for High-performance Computing} for providing computational resources and facilities.

{\small
\bibliographystyle{ieee}
\bibliography{normalization}
}

\end{document}